\documentclass[conference]{IEEEtran}
\IEEEoverridecommandlockouts

\usepackage{cite,subfigure,float}
\usepackage{amsmath,amssymb,amsfonts,bbm}
\usepackage{algorithmic}
\usepackage{graphicx}
\usepackage{textcomp}
\usepackage{tcolorbox}
\usepackage{tabularx}
\usepackage{xcolor}
\usepackage{amsmath}
\usepackage{multicol}

\usepackage{hyperref}
\hypersetup{draft}
\DeclareMathOperator*{\argmin}{arg\,min}
\def\BibTeX{{\rm B\kern-.05em{\sc i\kern-.025em b}\kern-.08em
    T\kern-.1667em\lower.7ex\hbox{E}\kern-.125emX}}
\newcommand*{\affaddr}[1]{#1} 
\newcommand*{\affmark}[1][*]{\textsuperscript{#1}}
\newcommand{\indicator}{\mathbbm{1}}
\def\uR{{\mathbb R}}

\begin{document}
\title{Federated Graph Learning for Low Probability of Detection in Wireless Ad-Hoc Networks\thanks{This research was supported by Defence Science and Technology’s Artificial Intelligence for Decision Making Initiative (AI4DM) program.}\\

}

\author{%
Sivaram Krishnan\affmark[1], Jihong Park\affmark[1],  Subhash Sagar\affmark[1], Gregory Sherman\affmark[2], Benjamin Campbell\affmark[2], and  Jinho Choi\affmark[1] \\
\affaddr{\affmark[1]School of Information Technology, Deakin University,  Australia }\\
\affaddr{\affmark[2]Defense Science and Technology Group, Australia}\\

}

\maketitle

\begin{abstract}
Low probability of detection (LPD) has recently emerged as a means to enhance the privacy and security of wireless networks. Unlike existing wireless security techniques, LPD measures aim to conceal the entire existence of wireless communication instead of safeguarding the information transmitted from users. Motivated by LPD communication, in this paper, we study a privacy-preserving and distributed framework based on graph neural networks to minimise the detectability of a wireless ad-hoc network as a whole and predict an optimal communication region for each node in the wireless network, allowing them to communicate while remaining undetected from external actors. We also demonstrate the effectiveness of the proposed method in terms of two performance measures, i.e., mean absolute error and median absolute error.
\end{abstract}


\section{Introduction}
In contemporary warfare, wireless connectivity among military equipment and personnel is essential for effective communication and coordination. In such a scenario, we need to ensure the entire communication (i.e., transmission) remains undetected while hiding the content of the transmission \cite{yan2019low}. For example, imagine a scenario where a person (hereafter referred to as a node in the study) seeks to transmit information to a base station while avoiding detection by external actors who typically employ radio frequency (RF) signal detection techniques, such as electronic signal measures (ESMs), to detect, locate, and identify the sources of RF signals. Detection of RF signals poses a significant risk to transmission success and privacy of both sender and recipient \cite{zeng2011parameter}, thus LPD methods has elicited a significant amount of research interest in recent times.

Research on LPD methods addresses the issue of not only hiding the content of a communication, but also the entire transmission itself \cite{yan2019low}, \cite{yang2008low}. In such wireless communication operations, the optimisation of transmission power at each sending node is a major concern, which is addressed by studying LPD methods. While most works on LPD investigate the scenario of a single point-to-point transmission between two nodes in the presence of an external actor \cite{sobers2017covert}, \cite{bash2013limits}, there is limited research on scenarios of wireless ad-hoc networks with multiple nodes, as seen in Fig.~\ref{Fig:LPD_}. Furthermore, the limited studies of wireless ad-hoc networks often rely on iterative algorithms that are computationally expensive and often lack closed-form analysis \cite{sheikholeslami2018multi}.

This research uses a machine learning-based framework to address multiple point-to-point communication in a wireless network. In general, the main contributions of the paper are as follows:

\begin{figure}
\centering
\includegraphics[width=0.450\textwidth]{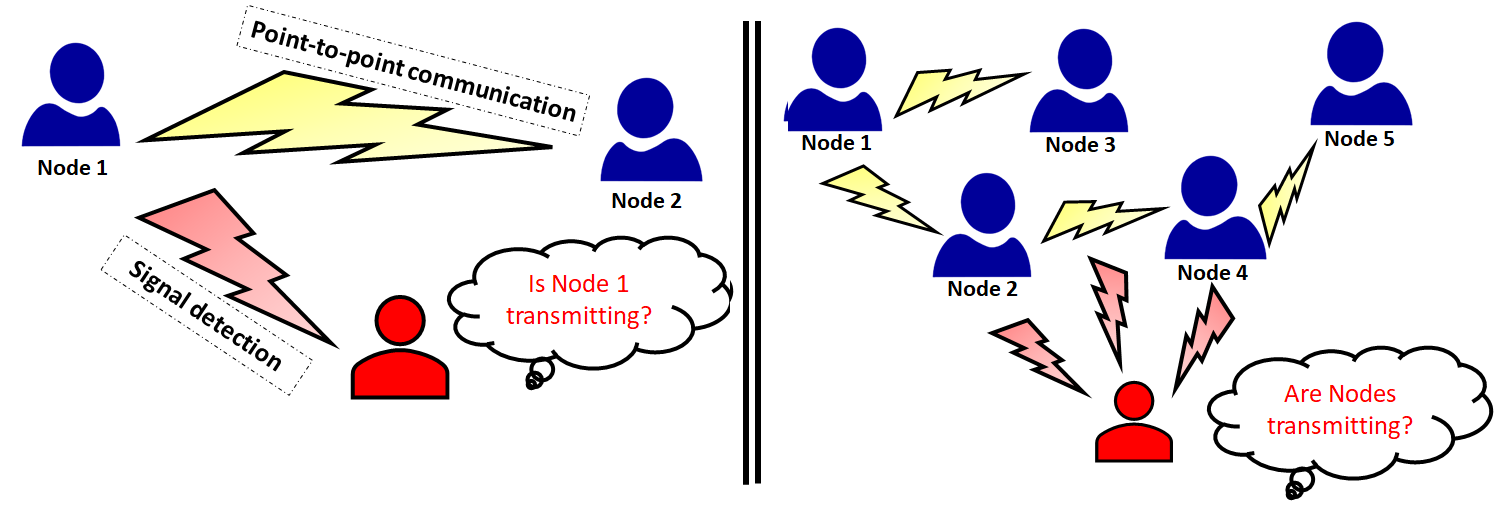}
\caption{A schematic illustration of point-to-point node communication (left) and multiple point-to-point
node communications (right).}
        \label{Fig:LPD_}
        \vspace{-1.35em}
\end{figure}
\begin{itemize}
\item To allow users in a wireless ad-hoc network to communicate with each other while minimising the chance of detection by an external actor for LPD, we propose a framework based on a graph neural network (GNN) with a snapshot of nodes' locations as the input so that the optimal communication area and connections for each node can be predicted \cite{he2021overview}, \cite{kipf2016semi}.

\item A federated learning (FL) model \cite{konevcny2016federated} is then proposed to generalise the capabilities of the GNN model and to address the challenge of limited data samples on a standalone architecture. In addition, a  pruning method is implemented to sparsify of the global model \cite{zhu2017prune}. 
\item Finally, the performance of the proposed framework is evaluated using mean absolute error (MAE) and median absolute error (MedAE). 
\end{itemize}
\section{System Model}
\label{sec:sys_mod}
Suppose there are $N$ nodes participating in a communication task where communication can take place through either direct links or multi-hop links. 

Let $u_{n}, n \in \{1, \cdots, N\}$, denote the locations of the nodes in the 2-dimensional space ($u_n \in \mathbb{R}^2$). Furthermore, each node is assigned with a coverage radius ($r_n \in \mathbb{R}^2$) as per its location in the network which determines it communication region.  The associated communication region or the coverage area of node $n$ is characterised by the circular disk as follows: 
\begin{equation} \label{range}
\mathcal{C}_n = \{x : ||u_n - x||_2 \leq r_n\}.  
\end{equation}

We aim to realise an LPD system for a wireless ad-hoc network where our primary objective is to minimise the communication region of the entire wireless network ($\bigcup_{n}\mathcal{C}_n$). This can be considered the minimum-area spanning tree (MAST) problem \cite{carmi2006minimum}, \cite{guimaraes2021minimum} and is also an approximation of the well-known minimum spanning tree (MST) problem \cite{graham1985history}. In the original MAST formulation, an undirected graph is considered where each node is associated with a circular disk, with the diameter of the disk corresponding to the length of the edge, which represents the distance between the communicating nodes. MAST seeks a spanning tree that minimises the union of the circular disks. Furthermore, utilising MAST as the foundation for our study, for the scenario in question, we have to take into account some additional constraints including but not limited to, a \emph{covertness constraint}, where communication must not be detected by any external actors and there needs to be a \emph{connectivity} guarantee among all the participating nodes, previously studied in \cite{campbell2020asynchronous}, \cite{campbell2018minimising}. 

In general, the MST is considered to approximate and resolve the MAST problem \cite{carmi2006minimum} due to its low computational complexity. However, this approach leads to a decline in the overall accuracy. In contrast, this research employs machine learning-based methods to address the MAST problem while maintaining high accuracy. The proposed machine learning-based model considers the following key assumptions.

\begin{itemize}
\item[{\bf A1}:] 

The proposed framework prioritises anonymity in communication over transmission power ($\mathcal{P}$), as a high transmission power increases the risk of detection \cite{sheikholeslami2018multi}. We assume that the information shared is small in size, so even with limited transmission power and a limited signal-to-noise ratio (SNR) at the receiver, the transmission will still be successful. The transmission power necessary for transmitting from node $i$ to node $j$ can be expressed as:
\begin{align}
\mathcal{P}_i = SNR(d_{ij})  N_0   d_{ij}^{\eta}, 
\  \forall i, j \in \{1, \cdots, N\},
    \label{EQ:SNR}
\end{align}
where 
$d_{ij}$ denotes the distance between nodes $i$ and $j$, 
$SNR(d_{ij})$ is the SNR at node $j$, 
$N_0$ is the noise density, and
$\eta$ represents the path-loss exponent. In \eqref{EQ:SNR}, we ignore the small-scale fading as the information sequence can be sufficiently long for low-rate communication to average out small-scale fading.

\item[{\bf A2}:] It is also assumed that in order for link $l_{ij}$ to be established between nodes $i$ and $j$, the distance between the nodes, $d_{ij}$, needs to be lower than their coverage radius and can be given by
\begin{align}
l_{ij}= \indicator(d_{ik} \le  \min\{r_i, r_j\}),
\end{align}
where $\indicator(\cdot)$ is the indicator function.

\end{itemize}
\section{Federated Graph Learning for LPD: Architectures and Operations}
\label{sec:pf}

\subsection{Problem Formulation}
The proposed formulation can be expressed as an optimisation problem with a goal to design an LPD system. Let $\mathcal{A}$ represent the total communication area of a wireless network, which is the primary objective function. That is, by considering the aforementioned constraint from Sec.~\ref{sec:sys_mod}, the following coverage area is to be minimised:
\begin{align}
\mathcal{A} = \bigl|\bigcup_{n=1}^N \mathcal{C}_n \bigl|,
\end{align}
where $|\mathcal{C}|$ represents the area of $\mathcal{C} \in \uR^2$.
Then, the optimal coverage radius for each node can be determined by minimising the communication area for the entire network when considering the locations of the nodes by solving the following problem:
\begin{align}
\{r_1^{\ast}, \cdots r_N^{\ast}\} &= \argmin_{u_1, \cdots, u_N} \mathcal{A} \cr
\mbox{subject to} &  \ |l_n| \geq 1 ,
\label{eq:MAST}
\end{align}
where $l_n = \{i:\ l_{ni} = 1\}$ represents the set of direct links for node n.
Here, note that for a node to be a part of the network, it is necessary that the node be directly linked to at least one more node, i.e., $|l_n| \ge 1$.
\begin{figure*}[ht]
    \centering
    \includegraphics[width=0.57\textwidth]{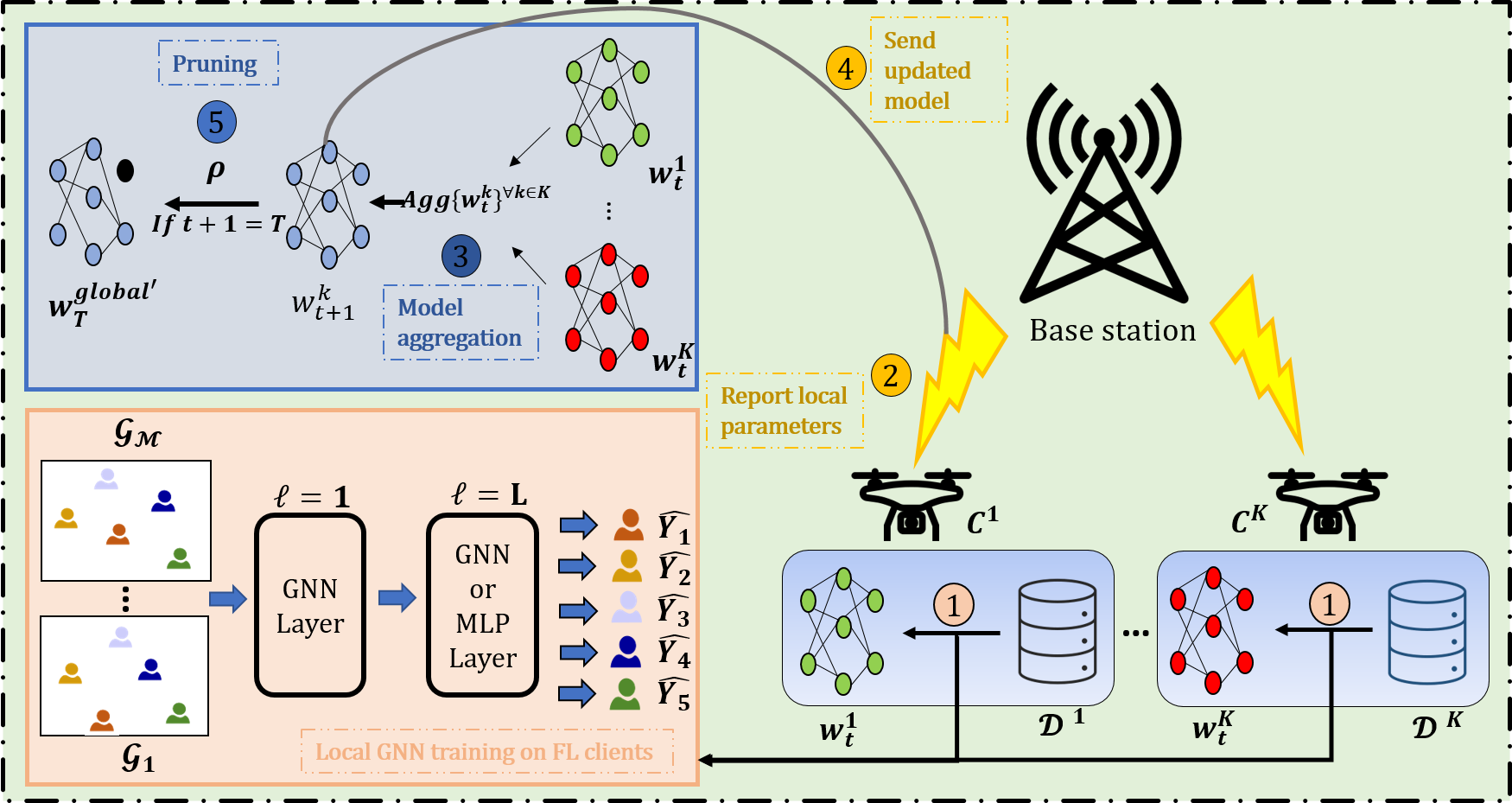}
    \caption{A schematic illustration of the proposed architecture.}
    \label{fig:img1}
    \vspace{-1.25em}
\end{figure*}

\subsection{GNN Architectures for LPD}

To solve the optimisation problem in \eqref{eq:MAST}, we utilise a GNN that aims to predict a communication region for each node in the graph (i.e., node regression) by feeding the locations of nodes (i.e., network topologies) as input samples. The graph convolution network (GCN) is the de facto standard GNN architecture consisting graph convolution layers \cite{kipf2016semi}. At the $\ell$th layer with weights $\boldsymbol{w}^{(\ell)}$, for the $i$th node with a set $\mathcal{N}_i$ of neighbours, a graph convolution layer yields the output activation $\boldsymbol{h}_i^{(\ell+1)}$ as follows:
\begin{align}
\text{(GCN)}\quad & \boldsymbol{h}_i^{(\ell+1)} = \sigma \Bigg( \sum_{j\in \mathcal{N}_i} \frac{1}{c_{ij}} \boldsymbol{w}^{(\ell)} \boldsymbol{h}_j^{(\ell)} \Bigg),
\end{align}
where $\sigma(\cdot)$ is the rectified linear unit (ReLU) activation, and $c_{ij}$ is a constant that is pre-determined by the adjacency matrix of the input graph. 

Graph convolution layers are designed for processing rich features per each node, whereas in our LPD scenario, the feature set associated with each node is limited, under which training often fails to converge in our experiments. To resolve this issue, inspired from the graph attention network (GAT) \cite{velivckovic2017graph}, we additionally apply graph attention layers. At the $\ell$th layer for the $i$th node, an attention layer yields the following output activation:
\begin{align}
\text{(GAT)}\quad & \boldsymbol{h}_i^{(\ell+1)} = \sigma \Bigg( \sum_{j\in \mathcal{N}_i} \alpha_{ij} \boldsymbol{w}^{(\ell)} \boldsymbol{h}_j^{(\ell)} \Bigg),
\end{align}
where $\alpha_{ij}= \text{Softmax}(\text{Att}(\boldsymbol{h}_i, \boldsymbol{h}_j))$. and $\text{Softmax}(\cdot)$ and $\text{Att}(\cdot)$ are the softmax and dot-product functions, respectively. As opposed to the fixed $c_{ij}$ in graph convolution layers, $\alpha_{ij}$ additionally provides the trainable attention value between the nodes $i$ and $j$, thereby compensating sparse features in the LPD scenario. In Sec.~\ref{Sec:Sim}, we will demonstrate the effectiveness of our proposed GCN+GAT architecture for LPD, as compared with other baselines including the standard GCN, as well as the convolutional neural network (CNN) dedicated to processing images, which is a special case of a graph where the nodes (pixels) are connected only with neighbouring pixels.
\subsection{Federated GNN Training with Pruning}

To train a GNN for LPD, it requires a non-negligible amount of training samples that are unlikely to be co-located. In other words, the GNN training necessitates background communication to exchange network topology data. This however, can result in increased communication overhead, making it a naive solution with several drawbacks. To obviate this problem, we adopt FL for collaborative decision-making by exchanging model parameters during local training while maintaining local storage of data \cite{he2021fedgraphnn,konevcny2016federated}. Furthermore, to reduce the background communication cost that is proportional to the number of local model parameters \cite{kairouz2021advances}, we apply a model pruning method that removes redundant model parameters while maintaining prediction accuracy \cite{park2019wireless}. Fig. \ref{fig:img1} summarises the proposed framework consisting of the following three key operations: 1) graph-based learning (step 1) for LPD and 2) distributed graph learning using FL (stpng 2-4) and 3) weight pruning (step 5), as elaborated next. 

\subsubsection{Graph-based learning for LPD}

Formally, a graph is defined as $\mathcal{G} = (V, \mathcal{E}, \mathcal{X})$, where $\mathcal{V} = \{v_n\}$ defines the set of the vertices of a graph, which represents the nodes performing a communication task, $\mathcal{E} = \{(e_{ij})\}_{i,j \in \mathcal{V}}$ defines a set of edges connecting the nodes, representing the links between the nodes. In addition to the structure of a graph, the framework utilises the spatial information of the nodes in the form of additional node features $\mathcal{X} = \{x_n\}$. The function $f$ performs a regression task that maps the set of nodes $\mathcal{V}$ to the set of communication ranges as $f$: $\mathcal{V}$ $\rightarrow$ $\{Y_n\}$. Our proposed GNN approach, denoted by $\phi$, is trained to make predictions, denoted by denoted by $\{\hat Y_n\} = \ \phi$ \ ($\mathcal{G};w$), on unseen graphs.  We aim to find the optimal weights $w^{\ast}$ by minimising the loss function $\mathcal{L}_{MAE}$ over an input~$\mathcal{G}$:
\begin{equation}
w^{\ast} = \argmin_{\mathcal{G}} \underbrace{\frac{1}{N}\sum_{n=1}^{N}|Y_n-\hat{Y}_n| }_{:=\mathcal{L}_{MAE}}. \label{eq:loss}
\end{equation}

\subsubsection{Federated graph learning}
Each of the $K$ workers in the distributed learning scenario locally trains a GNN model $\{\phi^{k}\}^{\forall k = 1, \cdots K}$. The local dataset for these workers is denoted by $\mathcal{D}^k = \{(\mathcal{G}^{(k)}_m)\}_{{\forall m = 1, \cdots, \mathcal{M}}}$. Then each worker is optimised locally over local dataset $\mathcal{D}^k$ with $\mathcal{M}$ graphs, as:
\begin{equation}
w_t^{k^{\ast}} = \argmin_{w_t^{k}} \frac{1}{|\mathcal{M}|} \sum_{m = 1}^{|\mathcal{M}|}
\sum_{n = 1}^{N} |Y_{n}^{m} - \hat{Y}_{n}^{m}|,
\end{equation}
where $w_t^k$ denotes the global weights at the start of training round $t$. Upon successful completion of a local training at round $t$, the optimal weight $w_t^{k^{\ast}}$ is sent back to the global server for aggregation. In general, the global optimisation problem for federated learning can be written as:
\begin{equation}
w_{t}^{global^{\ast}} = \argmin_{ w_t^{k}} \frac{1}{K} \frac{1} {|\mathcal{M}|} \sum_{k = 1}^{K} \sum_{m = 1}^{|\mathcal{M}|}
\sum_{n = 1}^{N} |Y_{n}^{k,m} - \hat{Y}_{n}^{k,m}|,
\end{equation}
where $w_{t}^{global^{\ast}}$ represents the aggregated weight obtained after each communication round. The aggregated weight is then sent back to the local workers for the next training round, and this process continues for a fixed number $T$ of rounds. 
\subsubsection{Weight pruning}
A magnitude-based pruning method is applied to the final set of weights to sparsify the memory of the global model $w^{global}$. The pruning process is controlled by a threshold parameter, $\rho$, which determines the level of sparsity in the final set of weights. Additionally, strategically pruning allows the final model to be deployed on storage-constrained IoT devices for real-time predictions based on data collection.
\begin{align}
{w^{global'}} & = (1 - \rho)|w^{global}|  \cr
\mbox{subject to} & 
\ \Delta\mathcal{L}_{MAE}(\phi, \phi') \le \theta, \  \rho \in \{0, 1\},
\end{align}
where $\theta$ denotes a user-defined performance  loss threshold which must be greater than the difference of the loss function for predictions using the original GNN architecture $\phi$ with final weights $w^{global}$ given by $\mathcal{L}_{MAE}(\phi)$ and the loss function for the pruned GNN architecture $\phi'$ with pruned weights $w^{global'}$ given by $\mathcal{L}_{MAE}(\phi')$. 
\begin{figure*}
    \begin{centering}
    \subfigure[MLP vs. CNN vs. GCN.]{
    \label{fig:rob_acc_adv}
     \includegraphics[width=0.198\linewidth]{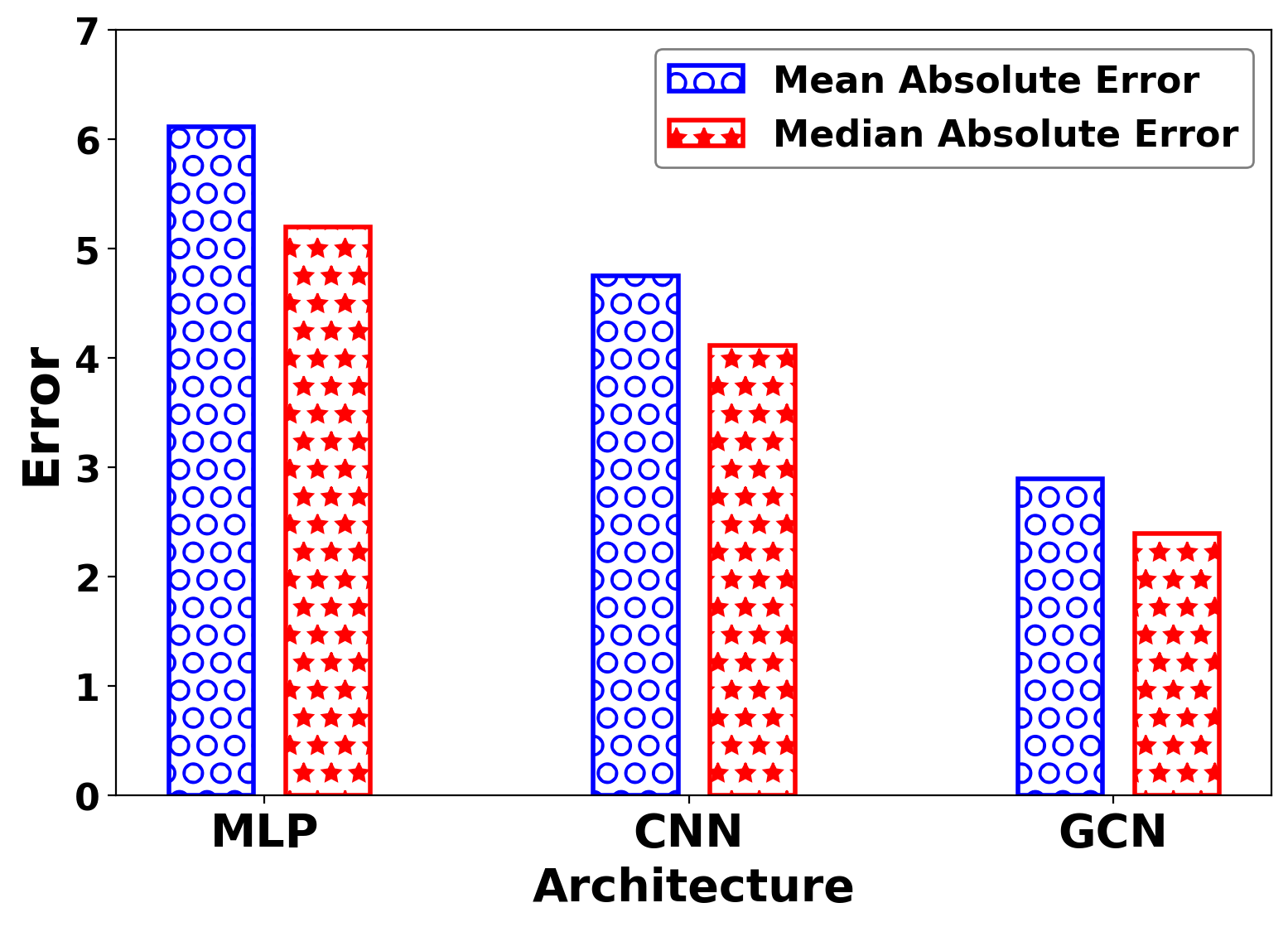}}\hfill
    \subfigure[Architecture with varying GNN + MLP layers]{\label{fig:rob_acc_bdv}\includegraphics[width=0.20\linewidth]{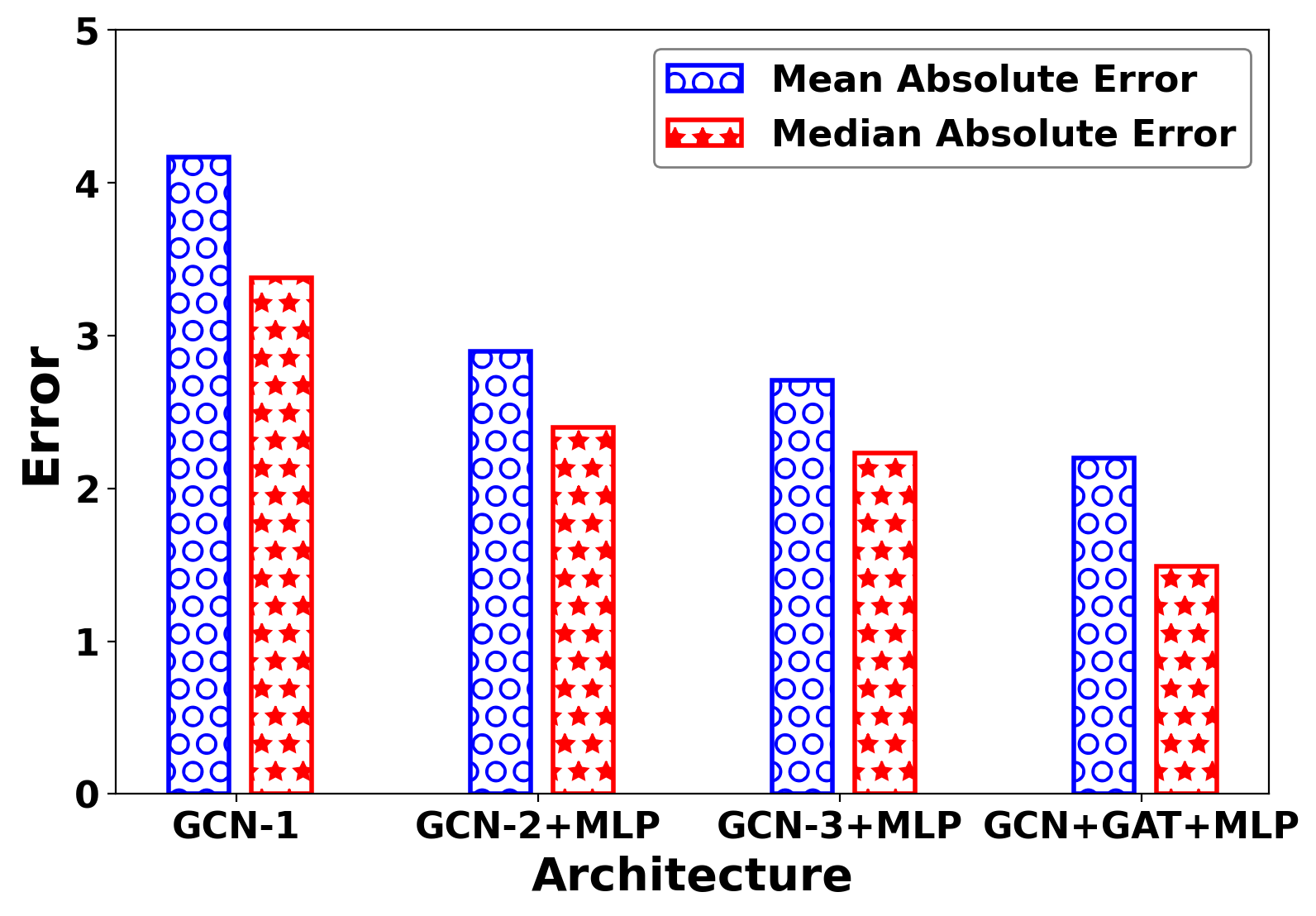}}\hfill
    \subfigure[Impact of local dataset sizes]{\label{fig:rob_acc_cdv}
    \includegraphics[width=0.20\linewidth]{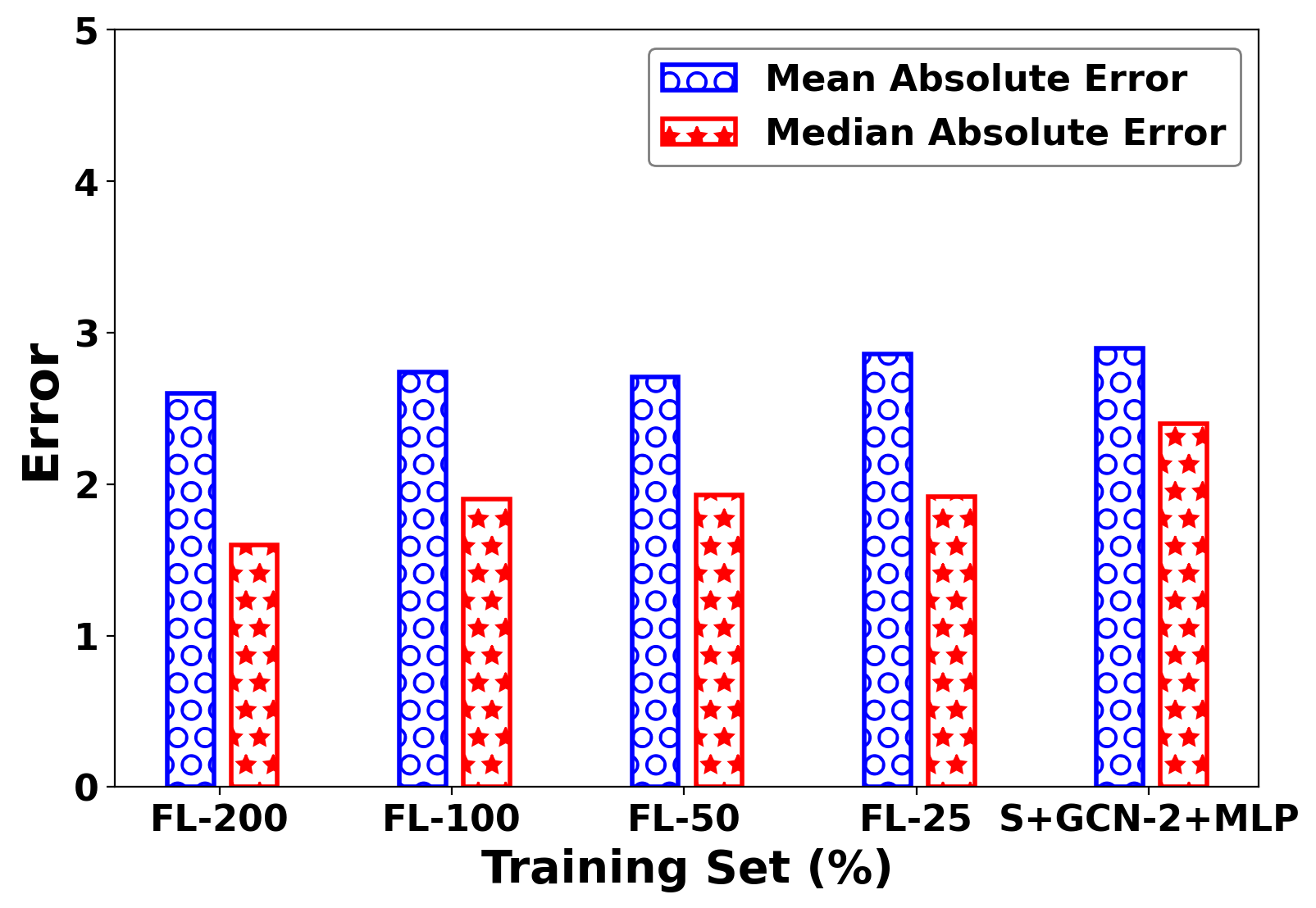}}\hfill
    \subfigure[Impact of sparsity levels]{
    \label{fig:rob_acc_ddv}
    \includegraphics[width=0.22\linewidth]{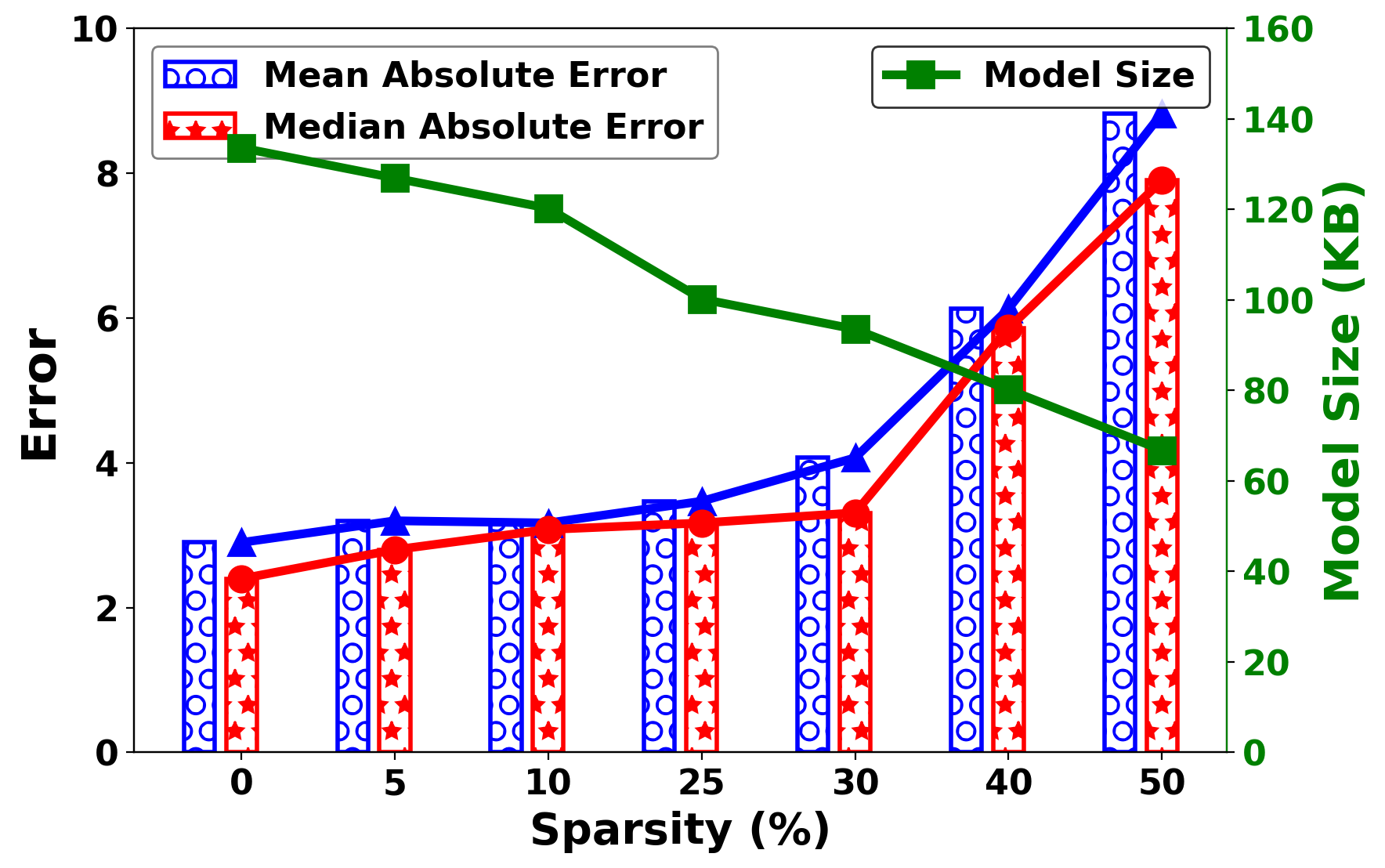}}
    \caption{Performance evaluation using MAE and MedAE}
    \end{centering}
\end{figure*}
\section{Experimental Setup and Results} \label{Sec:Sim}
\subsection{Experimental Setup}
For the experimental setup, we have generated a synthetic dataset of 200 graphs with 5 nodes each (i.e.,  vertices  in a graph). The nodes were randomly placed within a designated area of operation using a random point configuration of the Euclidean plane with each graph representing a distinct spatial layout. Realistically, the location of an external actor is unknown and as such has not been considered in our graphs. 

For the model training, $80\%$ of the dataset is utilised for training, and the remaining data is used for testing purposes.  The neural network architecture is trained with $\epsilon = 1000$ epochs and for the federated learning architecture, a total of 6 participant workers are selected with each trained for $\tau = 150$ rounds. 
\subsection{Results and Analysis}
The performance of the proposed framework is carried out by comparing the basic neural network architectures (i.e., multi-layer perception classifier (MLP) and CNN with a benchmark GNN architecture (i.e., GCN which consists of two graph convolution layers and a fully connected layer) - see Fig. \ref{fig:rob_acc_adv} with performance metrics, i.e., MAE, as seen in the Eq. \ref{eq:loss} and MedAE, to account for potential outliers in the loss values. It can be observed from Table \ref{table:tab1} that the MAE and MedAE score of GCN is lower with around $52.7\%$ and $58.7\%$ less than MLP, and $38.9\%$ and $41.7\%$ lower than CNN respectively. The reason for these results is the use of the graph convolution layer is designed to handle sparse and long-range communication through the implementation of a message-passing algorithm.  
\begin{table}[htb!]
\centering
\caption{MAE and MedAE score of employed architectures}
    \begin{tabular}{c|c|c}
    \hline
    \textbf{Architectures} & \textbf{Mean Absolute Error} & \textbf{Median Absolute Error} \\ \hline \hline
         MLP & 6.12 & 5.8 \\ \hline
         CNN & 5.8 & 4.12 \\ \hline
         GCN & \textbf{\underline{2.9}} & \textbf{\underline{2.4}} \\ \hline
         \hline
\end{tabular}
\label{table:tab1}
\end{table}

The GCN architecture can be improved further in order to be effective for inductive learning when dealing with limited input features \cite{xiao2022graph}. In Fig. \ref{fig:rob_acc_bdv}, we compared the performance of GCN architectures where GCN-$\ell$ refers to a GCN network architecture with $\ell$ graph convolution layers. We evaluated the performance of these GCN architectures against a hybrid GNN architecture that combines both graph convolution and graph attention layers. As can be seen in the figure, the performance of the variant GCN with MLP and GAT outperforms all the other variants of GCN, this is due to the self-attention mechanism employed in the GAT (using graph attention layer) that generates the hidden representation of each node. In essence, the significance of incorporating self-attention leads to an improved performance. 

As discussed in Sec.~\ref{sec:pf}, FL can be an effective solution for addressing the issue of limited sample size. For our study, it can be seen that FL improves the model performance as depicted in Fig. \ref{fig:rob_acc_cdv},. For instance, with FL-25 (where each worker with a sample dataset of 25 graphs) outperforms the standalone architecture having a single worker with a sample dataset of 180 graphs for training. In summary, the federated learning architecture FL-25 uses fewer data samples for training and still achieves a 20\% reduction in MedAE on unseen graphs. Additionally, the impact of varying sparsity levels on the performance of FL-25 is shown in Fig. \ref{fig:rob_acc_ddv} and it can be seen that while higher sparsity can lead to a higher error, the error remains relatively stable up until a sparsity level of 30\%, beyond which the error increases significantly. 
\section{Conclusion}
In this paper, we developed a GNN-based communication range (or equivalently transmit power) prediction framework for LPD. To reduce the training overhead without compromising security guarantees and communication efficiency, we additionally apply FL and model pruning during GNN training. Our proposed framework has been validated through simulation, demonstrating superior performance compared to baseline architectures in terms of prediction accuracy for unseen graph predictions and communication efficiency in FL when trained on limited data samples. Extending this study, encompassing realistic and dynamic scenarios under a 3-dimensional area of operation could be an interesting topic for future research.
\clearpage
\bibliographystyle{ieeetr}
\bibliography{refs}

\end{document}